\documentclass{article}

    \usepackage[preprint, nonatbib]{neurips_2019}

\usepackage[utf8]{inputenc} % allow utf-8 input
\usepackage[T1]{fontenc}    % use 8-bit T1 fonts
\usepackage{hyperref}       % hyperlinks
\usepackage{url}            % simple URL typesetting
\usepackage{booktabs}       % professional-quality tables
\usepackage{amsfonts}       % blackboard math symbols
\usepackage{nicefrac}       % compact symbols for 1/2, etc.
\usepackage{microtype}      % microtypography
\usepackage{graphicx}
\usepackage{xcolor}
\usepackage{caption}
\usepackage{subcaption}
\usepackage{wrapfig}
\usepackage{algorithm}% http://ctan.org/pkg/algorithms
\usepackage{algorithmic}% http://ctan.org/pkg/algorithms
\usepackage[title]{appendix}

\usepackage{amsmath,amsfonts,bm}

\def\eqref#1{equation~\ref{#1}}
\def\1{\bm{1}}

\DeclareMathAlphabet{\mathsfit}{\encodingdefault}{\sfdefault}{m}{sl}
\SetMathAlphabet{\mathsfit}{bold}{\encodingdefault}{\sfdefault}{bx}{n}

\newcommand{\KL}{D_{\mathrm{KL}}}

\DeclareMathOperator*{\argmax}{arg\,max}

\def\citet{\cite}
\def\citep{\cite}
\newcommand{\pluseq}{\mathrel{+}=}

\def\EE{\mathbb{E}}

\def\EE{\mathbb{E}}

\newcommand{\Ent}{\mathsf{H}}

\title{Meta Reinforcement Learning As Task Inference}

\author{%
  Jan Humplik \\
  DeepMind \\
  \And
  Alexandre Galashov \\
  DeepMind \\
  \And
  Leonard Hasenclever \\
  DeepMind \\
  \And
  Pedro A.\ Ortega \\
  DeepMind \\
  \And
  Yee Whye Teh \\
  DeepMind \\
  \And
  Nicolas Heess \\
  DeepMind \\
}

\begin{document}

\maketitle

\begin{abstract}
Humans achieve efficient learning by relying on prior knowledge about the structure of naturally occurring tasks. There is considerable interest in designing reinforcement learning (RL) algorithms with similar properties. This includes proposals to learn the learning algorithm itself, an idea also known as meta learning. One formal interpretation of this idea is as a partially observable multi-task RL problem in which task information is hidden from the agent. Such unknown task problems can be reduced to Markov decision processes (MDPs) by augmenting an agent's observations with an estimate of the belief about the task based on past experience. However estimating the belief state is intractable in most partially-observed MDPs. We propose a method that separately learns the policy and the task belief by taking advantage of various kinds of privileged information. Our approach can be very effective at solving standard meta-RL environments, as well as a complex continuous control environment with sparse rewards and requiring long-term memory.
\end{abstract}

\section{Introduction}\label{sec:introduction}

Recent advances in reinforcement learning (RL) combined with deep learning have led to rapid progress in difficult domains (e.g. \citet{mnih2015human, silver2017mastering, heess2017emergence, openai2018learning}). Remarkably, the RL algorithms which solved these tasks succeeded with surprisingly little access to prior knowledge about the task structure. Whilst the ability to learn with minimal prior knowledge is desirable, it can lead to computationally intensive learning.
This learning inefficiency  should be contrasted with human behavior. When trying to master a new skill, our learning progress relies heavily on prior knowledge that we have collected while solving previous instances of similar problems. The hope is that artificial agents can similarly
develop the ability to quickly learn if they have been previously trained in sufficiently rich multi-task settings in which the ability to learn is essential for success.

To study the emergence of efficient learning grounded in prior knowledge about a task distribution, recent papers have explored a  ``meta" perspective on reinforcement learning, e.g.\ \cite{duan2016rl2, wang2016learning,finn2017maml, mishra2018simple, ritter1018}. The standard approach here is to treat learning to solve a task as an optimisation problem, and to ``optimise the optimiser'' to speed up learning, i.e.\ learning-to-learn. 
An alternative perspective to meta-learning as optimisation is to think about the process of learning to solve a task as one of probabilistically inferring the task identity given observations \citep{feifei2003bayesian,finn2018probabilistic,garnelo2018conditional,garnelo2018neural,rakelly2019efficient}. Probabilistic inference makes sense in this scenario as we would like our agent to learn fast from small numbers of observations or interactions, and in this low information scenario there is necessarily uncertainty about the task identity. In a reinforcement learning setting, capturing this task uncertainty is important in order that the agent can properly balance exploration to identify the task and exploitation to maximise rewards given current knowledge of the task. Capturing uncertainty at the level of tasks can also aid smarter and more coherent exploration strategies, for example using heuristics like Thompson sampling (e.g. \citet{osband2017post}).

A natural approach to probabilistic meta reinforcement learning (meta-RL) is using a particularly formulated partially observed Markov decision processes (POMDPs). Suppose each task is described by a Markov decision process (MDP). Then an optimal agent which starts off not knowing the task is one that maximises rewards in a POMDP with a single unobserved (and static) state consisting of the task specification (e.g.\ reward function or transition probabilities) (see \cite{duff2002optimal, poupart2006analytic, brunskill2012bayes}). We will refer to this POMDP as a meta-RL POMDP (in Bayesian RL it is also called Bayes-adaptive MDP).

In general, the optimal policy in a POMDP is a function of the full history of actions, observations, and rewards. This dependence on the agent's experience can be captured by a sufficient statistic called the \emph{belief state} \citep{kaelbling1998planning}. In the case of the meta-RL POMDP, the relevant part of the belief state is the posterior distribution over the unknown task specification, given the agent's experience thus far. Reasoning about this belief state is at the heart of Bayesian reinforcement learning \citep{strens2000bayesian}, and many algorithms with optimal regret guarantees such as Thompson sampling \citep{agrawal2013further} effectively separate the algorithm for estimating the belief state from that for acting based on this estimate. 

In this work we exploit a similar separation of concerns between task inference and acting, in situations in which analytic solutions are intractable. In particular, we develop a two-stream architecture for meta-RL that augments the agent with a separate belief network whose role it is to estimate the belief state. 
This belief network is trained in a supervised manner using auxiliary, privileged, information available during meta-learning. Specifically, in typical meta-learning setups, the task distribution is under the designer's control, and task specifications are available at meta-training time as supervisory signal to train the belief network. Note that this privileged information is not required during meta-test time. 
Use of such auxiliary information is common in reinforcement learning.  Examples include the use of expert trajectories in imitation learning \citep{schaal1997learning}, use of natural language instructions \citep{hermann2017grounded}, and design of curricula allowing agents to gain simpler skills before more complex ones \citep{florensa2017reverse}. 
In our meta-RL setting the auxiliary information in the form of task specifications can be obtained essentially for free.

Our main contributions are: 
1) We demonstrate that leveraging this cheap task information during meta-training is a simple and cheap way to boost the performance of meta-RL agents; 
2) We show that we can train such meta-RL agents with recurrent policies efficiently using off-policy algorithms; 
3) We experimentally demonstrate that our agents can solve difficult meta-RL problems in continuous control environments, involving sparse rewards, and requiring smart search strategies and memorisation of information spanning more than 100 time steps.
4) We show that our agents discover the Bayes-optimal search strategy and we compare it to agents from meta-RL literature which rely on suboptimal strategies motivated by Thompson sampling.

\paragraph{Preliminaries}
Our method relies on basic results for MDPs and POMDPs. A MDP is a tuple~$(\mathcal{X}, \mathcal{A}, P, p_0, R, \gamma)$, where $\mathcal{X}$ is the state space, $\mathcal{A}$ is the action space, $P(x'|x, a)$ is the transition probability between states $x, x' \in \mathcal{X}$ due to an action $a \in \mathcal{A}$, $p_0(x)$ is the distribution of initial states, and $R(r|x, a, x')$ is the probability of obtaining reward $r$ after transitioning to a state $x'$ from $x$ due to an action $a$.
A POMDP is a tuple~$(\mathcal{X}, \mathcal{A}, P, p_0, R, \Omega, O, \gamma)$ which generalizes MDPs by including an observation space $\Omega$, and the probability $O(o'|x', a)$ of observing $o' \in \Omega$ after transitioning to a state $x'$ due to an action $a$. The states in POMDPs are not typically observed.
We denote sequences of states as $x_{0:t} = (x_0, \ldots, x_t)$, and similarly for observations, actions, and rewards. In POMDPs, we further define the observed trajectory as $\tau_{0:t} = (o_{0:t}, a_{0:t-1}, r_{0:t-1})$.
Given a policy $\pi(a_t|\tau_{0:t})$, the joint distribution between the states and the trajectory factorizes as $p_{\pi}(\tau_{0:t}, x_{0:t}) = p_0(x_0)O(o_0|x_0) \prod_{t'=0} ^ {t-1} \pi(a_{t'}|\tau_{0:t'})P(x_{t'+1}|x_{t'}, a_{t'})
O(o_{t'+1}|x_{t'+1}, a_{t'})R(r_{t'}|x_{t'}, a_{t'}, x_{t'+1})$.

The solution to a POMDP is a policy $\pi^*(a_t|\tau_{0:t})$ which maximizes discounted returns \citep{kaelbling1998planning}, i.e. $\pi^* = \argmax_{\pi} \EE_{\tau_{0:\infty}, x_{0:\infty} \sim p_{\pi}} \left[\sum_{t=0}^{\infty} \gamma^t r_t\right]$.
Note that conditioning the policy on past rewards is a subtle, yet important, generalization of what is typically assumed in POMDPs \citep{izadi2005using}. The optimal policy's dependence on the trajectory can be summarized using the so-called \emph{belief state} $b_t(x) \equiv p_\pi(x_t=x|\tau_{0:t}) 
%= \sum_{x_{0:t-1}} p_{\pi}(\tau_{0:t}, (x_{0:t-1}, x)) / \sum_{x_{0:t}} p_{\pi}(\tau_{0:t}, x_{0:t})
$, which is the conditional distribution over the state $x_t$ given trajectory $\tau_{0:t}$.
Note that the belief state $b_t$ is a deterministic function of the trajectory, and also that it is a sufficient statistic for the optimal $a_t$ in the sense that $\pi^*(a_t|\tau_{0:t}) = \pi^*(a_t|b_t)$ \citep{kaelbling1998planning}. However, in many tasks, it is not necessary to have access to the full belief state in order to act optimally.

\section{Meta reinforcement learning and task inference}
\label{sec:task_inference}

In this section we develop our approach which is motivated by viewing meta-RL as task inference. 
Let $\mathcal{W}$ be a space of tasks and $p(w)$ be a distribution over tasks $w\in\mathcal{W}$. 
We assume that each task $w$ is a MDP $(\mathcal{X}, \mathcal{A}, P^w, p_0^w, R^w, \gamma)$. All MDPs share the discount parameter and state and action spaces, while transition, initial and reward distributions are task specific. 
Our aim is to train an agent that maximizes future discounted rewards based on a small number of interactions with an unknown task $w$ drawn from $p(w)$. For simplicity of exposition we will assume that  interactions occur within a single episode; multi-episode interactions can be achieved by changing the MDPs to reset to the initial state, say every $T$ steps. 

We can formulate our meta-RL problem using a POMDP as follows. The POMDP shares the same action space $\mathcal{A}$ as the MDPs, and has states $(x, w) \in \mathcal{X} \times \mathcal{W}$, transition distribution $P(x', w'|x, w, a) = \delta(w' - w) P^w(x'|x, a)$, reward distribution $R(r|x, w, a, x', w') = R^w(r|x, a, x')$, initial state distribution $p_0(x, w) = p(w) p_0(x|w)$, and deterministic observations $O(o'|x', w, a) = \delta(o'-x')$.  In other words, the task specification is the only unobserved state and is drawn from $w\sim p(w)$ at the beginning and held fixed; subsequently the environment evolves according to task $w$'s MDP.

The optimal agent $\pi^*(a_t|\tau_{0:t})$ which does not have access to task labels $w$ but is assumed to have access to past observed interactions with the task, e.g.\ using a LSTM memory system, solves:
\begin{equation} \label{eq:main}
    \max_{\pi} \sum_{w} p(w) \sum_{\tau_{0:\infty}} p_{\pi}(\tau_{0:\infty}|w)\left[\sum_{t=0}^{\infty} \gamma^t r_t\right],
\end{equation}
where $p_{\pi}(\tau_{0:t}|w)$ is as given in Section \ref{sec:introduction} but for task $w$.

The belief state of this POMDP is $b_t(x, w) = p(x, w|\tau_{0:t}) = \delta(x - x_t)p(w|\tau_{0:t})$, where $p(w|\tau_{0:t})$ is the posterior over tasks given what the agent has observed so far. Crucially, because the agent itself has no access to $w$, we show in Appendix A that the posterior satisfies 
\begin{align}
    p(w|\tau_{0:t}) \propto
    p(w)p_0(x_0|w)\prod_{t'=0}^{t-1} P(x_{t'+1}|x_{t'},a_{t'},w)R(r_{t'}|x_{t'},a_{t'},x_{t'+1},w).
\end{align}

\begin{align}
    p(w|\tau_{0:t}) \propto
    p(w)p_0(x_0|w)\prod_{t'=0}^{t-1} P(r_{t'}, x_{t'+1}|x_{t'},a_{t'},w)
    \label{eq:posterior_independence}
\end{align} 

That is, given the trajectory $\tau_{0:t}$, the posterior is independent of the policy which generated the trajectory.
This result will allow us to derive an off-policy algorithm later. We overload notation and refer to the posterior alone as the belief state $b_t(w) = p(w|\tau_{0:t})$ since it is the only interesting part.  

As noted in Section \ref{sec:introduction} and detailed in Appendix A, the belief state is a sufficient statistic of the past. This has a natural interpretation in the meta-RL setup: when acting at time $t$, the optimal meta-learner only needs to make decisions based on the current state $x_t$ and the current belief $b_t(w)$ about which task it is solving. The implication being that we can restrict ourselves to policies of the form $\pi(a_t|\tau_{0:t}) \equiv \pi(a_t|x_t,b_t)$. 
In practice, the belief state is intractable to compute in all but the simplest POMDPs, and in any case computing it would require detailed knowledge of the required POMDP conditional distributions. However, this argument motivates an agent consisting of two modules: one being the policy $\pi(a_t|x_t,\hat{b}_t)$ dependent on the current state and (an approximate representation of) the posterior belief over tasks, and the other being a \emph{belief module} which learns to output an approximate representation $\hat{b}_t$ of the belief.  In the rest of this section we will describe agents of this form as well as algorithms to train them.

As a final remark, the above discussion can be extended to the general case where tasks are themselves POMDPs rather than MDPs, and this is the case in one of our environments (Sec.~\ref{sec:numpad}). If the tasks are POMDPs, then, in addition to $x_t$ and $b_t$, $\pi$ might need to have access to additional information about the past (i.e.\ the belief state of the task-specific POMDP). For this reason we will often model the policy $\pi(a_t|\tau_{0:t},\hat{b}_t)$ to depend on the full trajectory using an LSTM in addition to the belief.

\subsection{Learning the belief network}\label{sec:belief_network}

There is a rich literature on learning belief representations in the POMDP literature \cite{oord2018representation,igl2018deep}. Most have taken an unsupervised approach to learning these belief representations, as in general there is no access to additional useful information about the true underlying belief state. However, unsupervised  learning of belief representations is difficult and is in general unsolved.
Fortunately, in our meta-RL scenario, because the task distribution is typically designed by the researcher, there is information that can be used to learn good belief representations.  Specifically, during each episode, the system that trains the agent has a succinct specification of the task $w$, e.g.\ the location of the goal or the expected reward associated with picking up each object in the environment. In this paper we consider using such information to train the belief module via auxiliary supervised losses.  Note that once the belief module is learnt, this information is not needed during test time.

We view use of such privileged task information as similar in purpose as use of expert trajectories in imitation learning \cite{schaal1997learning}, natural language instructions \cite{hermann2017grounded} and designed curricula \cite{florensa2017reverse} to speed up learning of RL agents. In the following we will explore use of different types of task information with varying levels of privilege, to understand their effect on the learning process. In the following we denote by $h_t$ the task information, which can be time-varying and depend on both task description $w$ and trajectory $\tau_{0:t}$.  

\emph{Task description.} Here 
we simply train a belief module $b_\theta(h_t|\tau_{0:t})$ to directly predict the true task description $h_t=w$. 
\emph{Expert actions.} Suppose for each training task we have access to an expert agent $\pi^{w;e}(a_t|\tau_{0:t})$ trained only on that task. We can train the belief module to predict the action chosen by the expert, $h_t=a^{w;e}_t$.
\emph{Task embeddings.} In our experiments we have a fixed and finite number $K$ of tasks used for training. Indexing these by $\{1,2,\ldots,K\}$, we can train the belief module to predict the index $i^w$ of the task $w$, $h_t=i^w$.  Alternatively, if we have an embedding $F^w\in\mathbb{R}^d$ of each task, say obtained during a pre-training phase (see Appendix B), we can predict $h_t=F^w$. Both expert actions and task embeddings are weaker forms of privileged information since they just require the ability to index and access individual training tasks, e.g.\ for pre-training expert agents or task embeddings. 

We will train belief modules $b_\theta(h_t|\tau_{0:t})$ to predict the task information in a supervised way, by minimising an auxiliary log loss  $\mathbb{E}_{p(h_t|\tau{0:t})}[-\log b_\theta(h_t|\tau_{0:t})]$, where the target distribution $p(h_t|\tau_{0:t})$ is the posterior distribution over $h_t$ given the trajectory. This is equivalent to minimising the KL divergence $\mathbb{KL}(p(h_t|\tau{0:t})\| b_\theta(h_t|\tau_{0:t}))$,
and guarantees that if the belief module is flexible enough we can learn a close approximation to the true posterior $b_t$. Note that this is the reverse KL than is typical in variational inference.
While the posterior distribution is not tractably computable in general, samples from the posterior can be obtained for free in our meta-RL setup. This is because for each training episode the pair $(w,\tau_{0:t})$ is drawn from the joint distribution $p(w)p_\pi(\tau_{0:t}|w)$ so $w$ is an exact sample from the posterior $p(w|\tau_{0:t})$, c.f.\ amortized inference \cite{gershman2014amortized,paige2016inference}. This implies that $h_t$ is an exact posterior sample as well since it is a function of $w$ and $\tau_{0:t}$. As a further elaboration, since the posterior over $w$ is independent of the policy given the trajectory (c.f.\ \eqref{eq:posterior_independence}), the posterior over $h_t$ is as well. This means that the belief network can be trained using off-policy data generated by previous policies. In our experiments, we train the belief network alongside the policy using trajectories and task information $h_t$ stored in a replay buffer.

\subsection{Architectures and algorithms}
\label{sec:architectures}

\begin{figure}
    \centering
    \includegraphics[width=.9\linewidth]{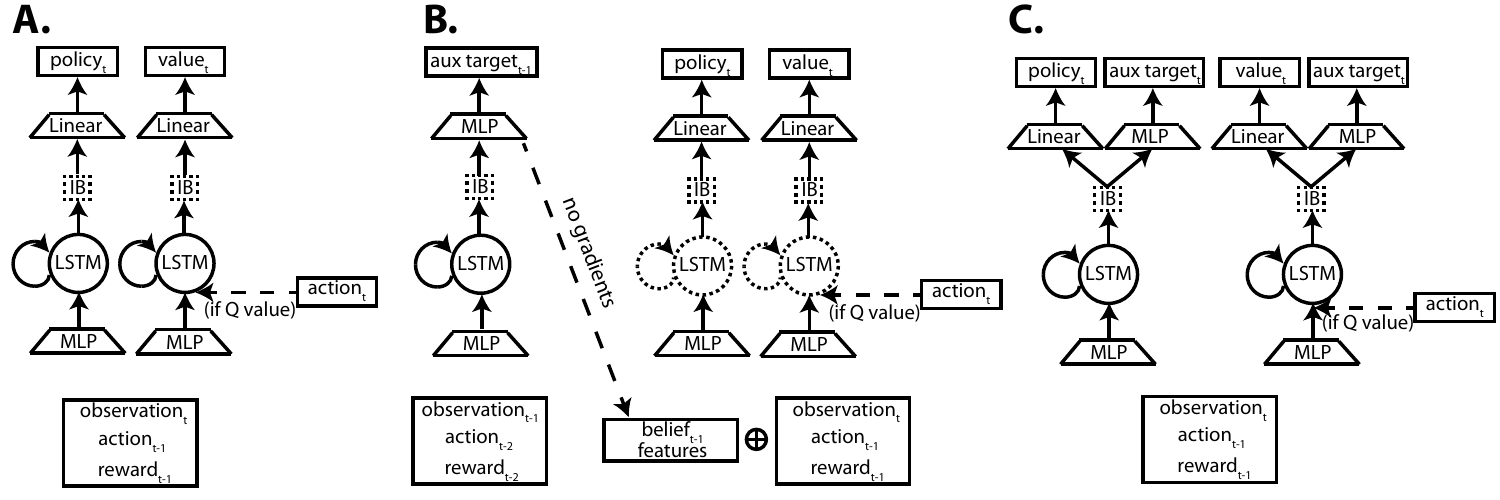}
    \caption{Agent architectures. A. Baseline LSTM agent. B. Belief network agent. C. Auxiliary head agent. LSTMs and information bottleneck (IB) are optional.}
    \label{fig:architectures}
\end{figure}

Fig.~\ref{fig:architectures}A
shows a baseline agent. It consists of two separate actor/policy and critic networks which embed inputs using a MLP before passing them to a LSTM \citep{hochreiter1997long}. This is similar to an architecture used, for example, in the $\text{RL}^2$ algorithm \citep{duan2016rl2}, and serves as a comparison point to recent meta-RL approaches. 
Fig.~\ref{fig:architectures}B describes our \emph{belief network} agent.
It augments the baseline agent with a belief network/module which outputs an approximate posterior over the auxiliary task information $h_t$. The actor and critic can either be recurrent with LSTMs as in the baseline agent, or feedforward with LSTMs omitted. Instead of the output layer, the actor and critic are fed a representation of the belief state given by the penultimate layer of the belief network. This is higher-dimensional and contains more information than the output layer, which can be quite low-dimensional (e.g.\ if task description is a 2D location of goal) or specific to training tasks (e.g.\ if using task index). Further the information in the final layer can be easily computed from the penultimate layer by the actor and critic networks if necessary.
In initial experiments (not reported here) we considered using samples from the learned posterior but this performed worse.
We do not backpropagate gradients from the actor and critic to the belief network. This ensures that the belief network focuses on learning a good representation of the belief state. It also means that we do not need to tune the balance between the RL and supervised belief prediction objectives, and interference between gradients of competing losses is not an issue.
Finally, in cases where $h_t$ is vector valued, we approximate the posterior using a factorized distribution which is expected to overestimate uncertainty. 
To demonstrate the advantage of our proposed architecture for the belief network agent, we also compared to an \emph{auxiliary head agent} in which the auxiliary supervised belief loss directly shapes the representations inside the actor and critic networks (see Fig.~\ref{fig:architectures}C). Detailed description of our architectures is in Appendix E and F.

In all three architectures, we explored regularizing the learnt representation using an information bottleneck (IB) \citep{alemi2017deep, chalk2016}. As shown in Fig.~\ref{fig:architectures}, we add a stochastic layer on top of the LSTMs in the actor, critic, and belief networks, and we regularize the noise in this layer by adding KL terms to each of these networks' losses. Details on IB are provided in Appendix C and E.

We mainly focus on off-policy training for which we use the SVG(0) algorithm \citep{heess2015learning} with an entropy regularization term in the policy update \citep{williams1991function,riedmiller2018learning}, however we will also compare to the on-policy PPO algorithm \citep{schulman2017proximal}. In experiments where we do not compare to PPO, we use a distributed version of SVG(0) which utilizes the Retrace operator \citep{munos2016retrace} for learning the action-value function. The supervised learning loss from Sec.~\ref{sec:belief_network} is optimized together with the policy and value losses on every iteration. We use the same distributed setup and strategy for initializing recurrent networks during off-policy learning as in \citet{liu2018emergent}. Hyperparameters and additional details are in Appendix E and F.

\subsection{Related works}

Meta reinforcement learning in which the goal is to learn an algorithm for quickly solving tasks drawn from some pre-specified distribution has recently received considerable attention. There are a few classes based on different architectures and inductive biases incorporated into the meta learner. Meta learners based on gradients typically utilize the MAML framework \citep{finn2017maml, gupta2018meta}. Another class uses RNNs to learn an internal representation of tasks \citep{duan2016rl2, wang2016learning, mishra2018simple, ritter1018}. 
%These approaches are closely related to learning to navigate in partially observable domains \citep{mirowski2016learning}.  
Several works have considered meta learning in a probabilistic setting where learning corresponds to inferring a task in a learnt hierarchical Bayesian model, e.g.\ \citet{feifei2003bayesian, garnelo2018neural, garnelo2018conditional}. This idea has recently led to alternative interpretations of ``neural'' approaches such as MAML \citep{grant2018recasting, yoon2018bayesian}. Our approach relates to both probabilistic and RNN approaches, but focuses on learning a good task representation using privileged information.
Meta RL which relies on privileged access to expert policies has also recently been studied in \citet{mendonca2019guided}.

In concurrent work, \citet{rakelly2019efficient} consider off-policy meta learning by separating control and task inference by also relying on privileged information about the training task ID. Their architecture is complementary to our IB regularized baseline agent in which the LSTM is replaced with a memory module that imposes a permutation invariant structure on its inputs which is justified in fully observed MDPs but cannot be easily extended to partially observed tasks. Furthermore, they force their agents to perform ``Thompson sampling"-like exploration which leads to suboptimal adaptation behavior.

Unsupervised learning of belief states to facilitate POMDP learning has been proposed in \citep{guo2019neural, moreno2018neural}. In a tabular setting, the usefulness of relying on the belief state for meta reinforcement learning of a small number of tasks has been proposed in \citep{brunskill2012bayes}.
More generally, our method is related to learning with auxiliary losses, which improve the performance of memory-based architectures both in reinforcement learning \citep{wayne2018unsupervised, jaderberg2016reinforcement, ke2019learning} and in sequence modelling \citep{trinh2018learning}. 
It is also related to reinforcement learning approaches which take advantage of the knowledge that the reward function or the environment have an additional structure related to multiple tasks, e.g.\ \citet{teh2017distral, wilson2007multi}. 

\section{Experiments}

Our experiments focus on demonstrating that: 
A. 
Direct training of the belief network with privileged information can speed up learning and lead to better final performance across a wide range of environments
B. 
We can train our recurrent agents efficiently off-policy; 
C. 
Information bottleneck regularization is an effective way for speeding up off-policy learning;
D. 
Our approach scales to a complex continuous control environment (Numpad) with sparse rewards and requiring long-term memory.

We ran experiments on environments of varying difficulty including common meta-RL testbeds. We train on 100 tasks and evaluate on a holdout set of 1000 tasks. We investigate generalization across tasks in Appendix I where we show that it is typically possible to generalize from 100 tasks. Detailed descriptions of environments are in Appendix G.

Most meta-RL papers consider a setup in which for each task the agent is given $K$ episodes of length $T$ to explore and adapt. Its performance is then measured either as the accumulation of rewards during these $K$ episodes \citep{duan2016rl2}, or as the rewards in new episodes after these $K$ ``adaptation'' episodes \citep{finn2017maml}. In the following we will be reporting the former measure to emphasize the exploration-exploitation trade-off inherent in meta-RL. 
We compare our \textbf{Belief} agents against \textbf{Baseline} LSTM and auxiliary head (\textbf{AuxHead}) agents (Fig.~\ref{fig:architectures}). Information bottleneck layers (\textbf{IB}) are included in all SVG(0) agents except otherwise noted. PPO agents do not utilize IB. Belief agents use LSTMs in all networks and learn using \textbf{task descriptions} except otherwise noted (e.g.\ \textbf{MDP}, \textbf{task id}). The Baseline agent is essentially $\text{RL}^2$, except that \citet{duan2016rl2} used TRPO while we used SVG(0) or PPO. As shown in \citet{rakelly2019efficient}, RL$^2$ with PPO is a stronger baseline than MAML, and so we omit comparisons to other meta-RL methods. 

\subsection{Off-policy and on-policy learning, and information bottleneck}

\begin{figure}
    \centering
    \begin{subfigure}[t]{0.49\linewidth}
        \vskip 0pt
        \centering
        \includegraphics[width=\linewidth]{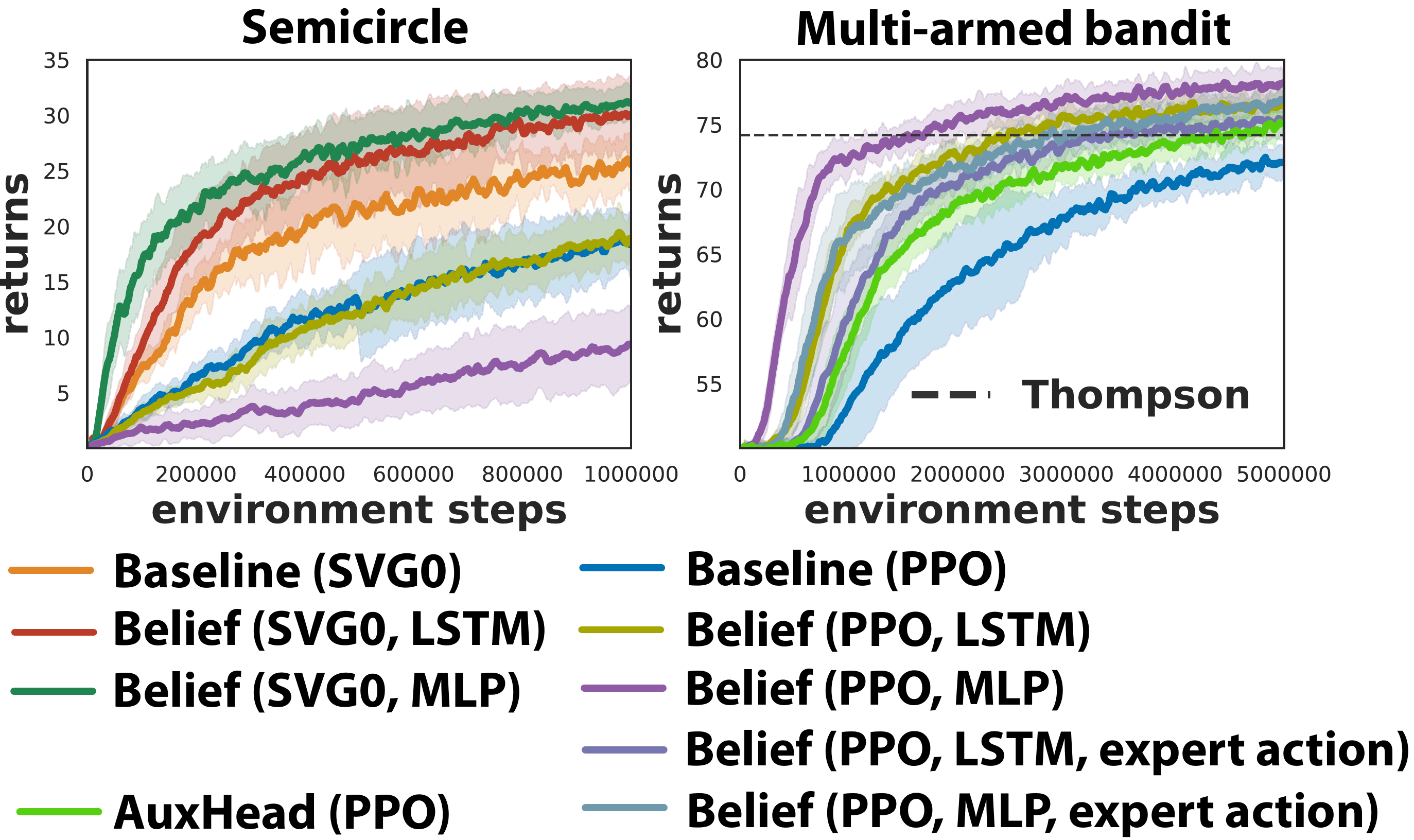}
        \caption{\label{fig:svg_vs_ppo}}
    \end{subfigure}
    \hfill
    \begin{subfigure}[t]{0.49\linewidth}
        \vskip 0pt
        \centering
        \includegraphics[width=\linewidth]{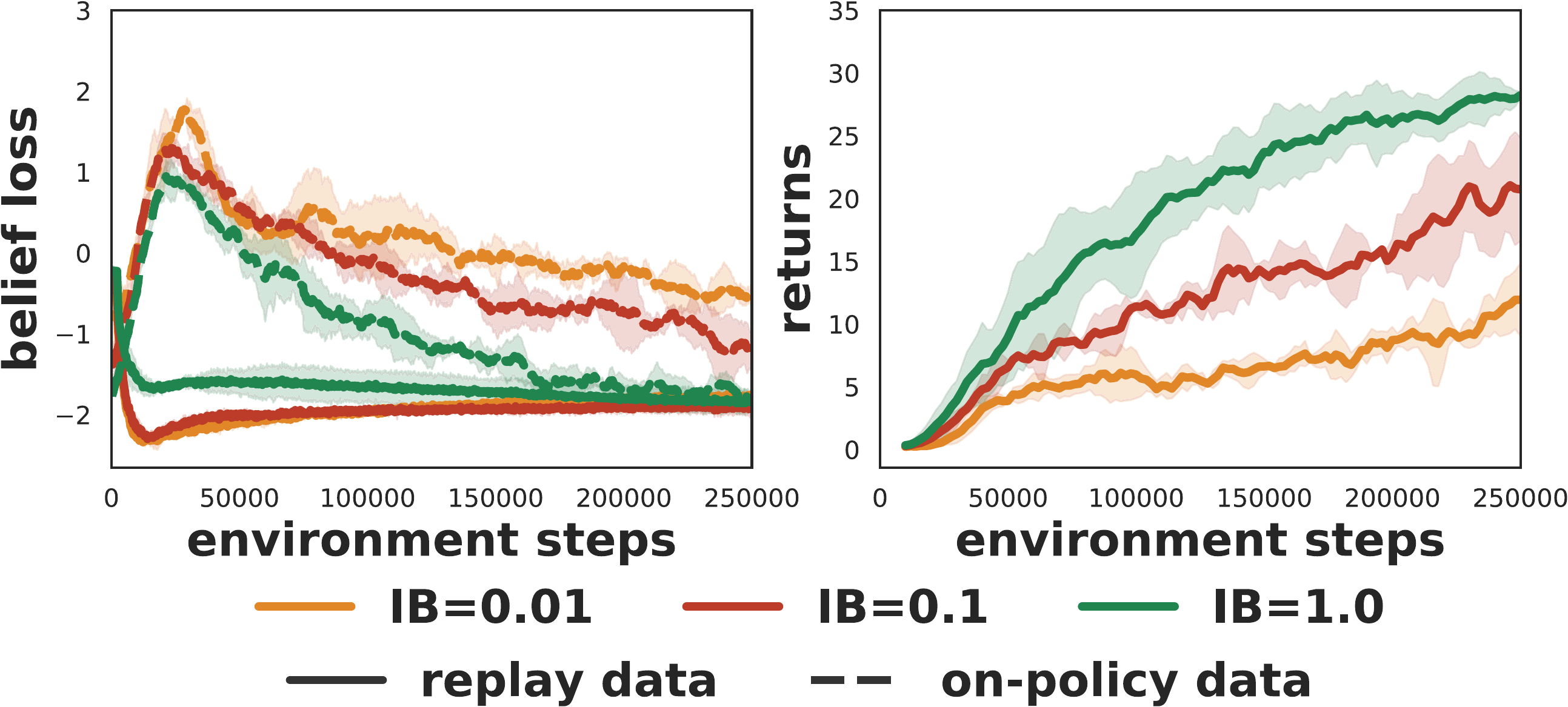}
        \caption{\label{fig:ib_analysis}}
    \end{subfigure}
    \caption{\textbf{(a)} Comparison of off-policy SVG(0) and on-policy PPO algorithms, and a demonstration of the advantage of supervision. \textbf{(b)} Increased IB regularization strength in the belief network leads to smaller generalization gap (left) and better performance (right) in the search for target task.}
    \label{fig:standard_results}
\end{figure}

We first compare an off-policy SVG(0) implementation to on-policy PPO, and investigate the importance of IB, on two simple environments studied in \cite{finn2017maml, duan2016rl2}: \textbf{Multi-armed bandit} with 20 arms and horizon 100, and \textbf{Semicircle} in which a point mass has to find a target on a semicircle. Only PPO is run on Bandit since it has a discrete action space.
Results are in Fig.~\ref{fig:standard_results}. In bandit, we also report performance of Thompson sampling.
Fig.~\ref{fig:svg_vs_ppo} shows that off-policy SVG(0) is more sample efficient than on-policy PPO with LSTM (which is similar to to RL$^2$ \citep{duan2016rl2}).
Belief agents generally perform better than Baseline agents. This is verified on harder environments below.
PPO on Semicircle is an exception due to the lack of off-policy replay data to train the belief network efficiently. 
For bandit, on policy data is sufficient. MLP agents, which do not use recurrent memories, can work only if the belief state is quickly effective.
In Bandit, we also trained the belief network using the expert action (best arm). This performed worse than task description (arm reward probabilities) but still better than the LSTM Baseline. AuxHead agents perform worse than the Belief agents.

The distribution of trajectories stored in a replay buffer during off-policy training is different from the distribution of trajectories generated by the current policy.
We found IB regularization to be a crucial component enabling this generalization. This is highlighted in Fig.~\ref{fig:ib_analysis}.  Fig.~\ref{fig:ib_analysis} (left) compares the belief network loss (negative log-likelihood) evaluated on the replay distribution and on the on-policy distribution in the semicircle task for several strengths of IB regularization. Fig.~\ref{fig:ib_analysis} (right) shows the corresponding learning curves. Increasing the regularization strength decreases the generalization gap, and it increases the sample efficiency of the agent.

\subsection{Further evaluation of the role of supervision}
\label{sec:further_results}
\begin{figure}
    % \centering
    \centering
    \includegraphics[width=.9\linewidth]{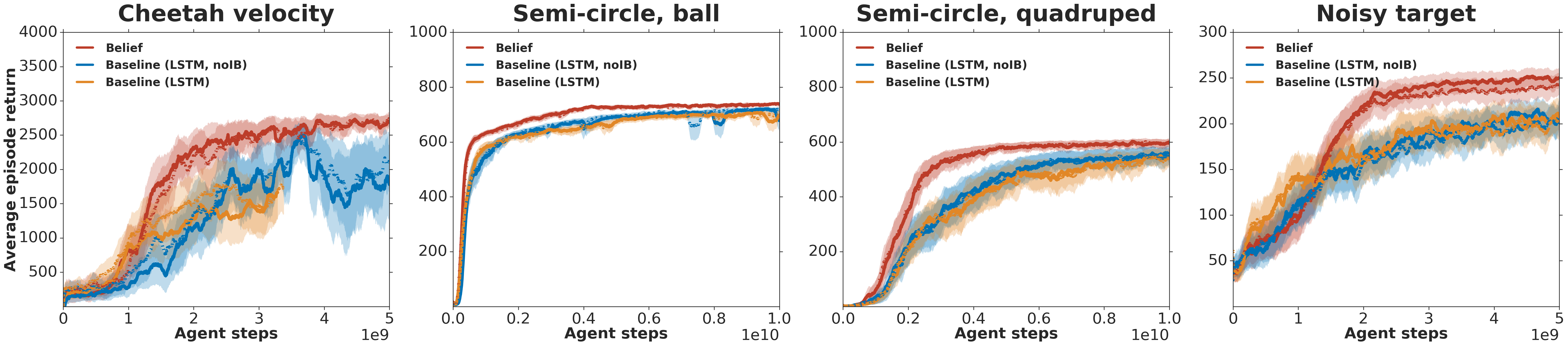}
    \caption{
    A summary of the advantage of using the belief network supervised with task descriptions. Solid curves correspond to performance on training tasks, while dashed curves to performance on validation tasks.}
    \label{fig:further_results}
\end{figure}

\begin{figure}
    \centering
    \hfill\hfill
    \begin{subfigure}[b]{0.27\linewidth}
        \centering
        \includegraphics[width=\linewidth]{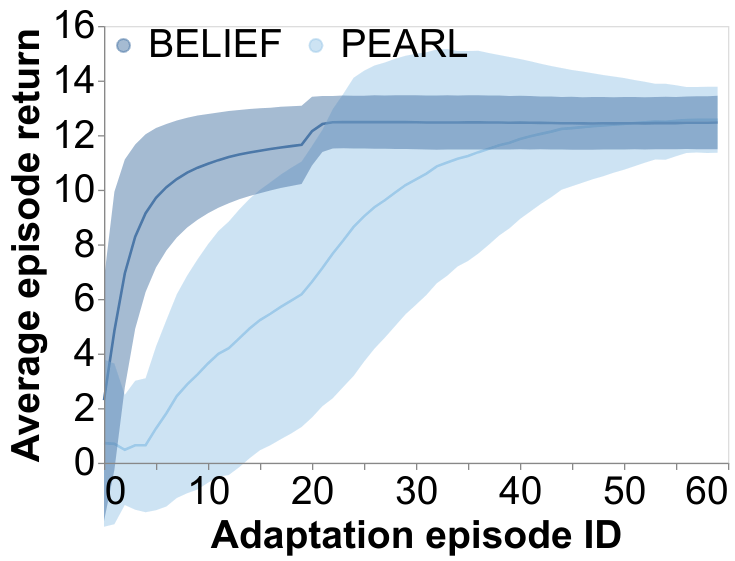}
        \caption{}
        \label{fig:pearl_vs_belief_adaptation}
    \end{subfigure}
    \hfill
    \begin{subfigure}[b]{0.54\linewidth}
        \centering
        \includegraphics[width=\linewidth]{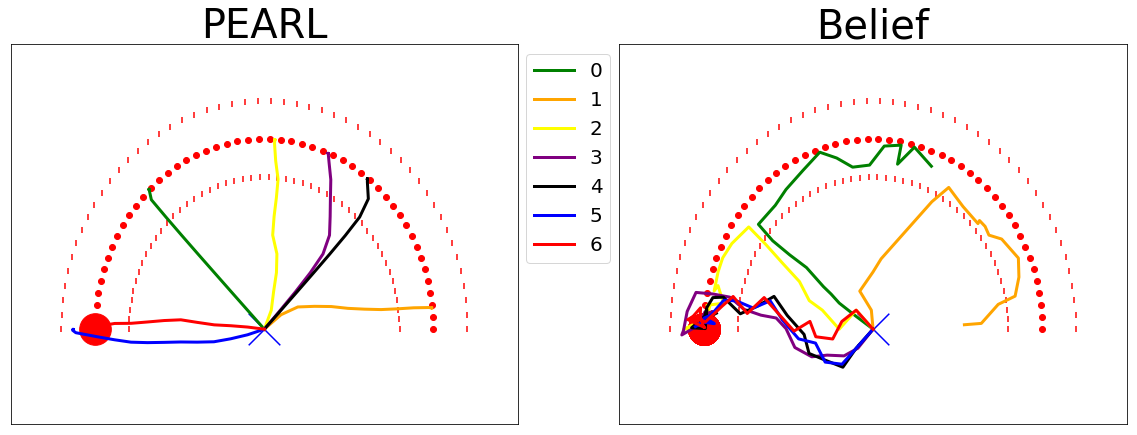}
        \caption{\label{fig:pearl_gts_trajectories}}
    \end{subfigure}
    \hfill\hfill
    \caption{\textbf{(a)} Comparison of the average episodic returns during adaptation of our Belief agent and PEARL agent in the sparse 2d navigation task from \cite{rakelly2019efficient}. The error bars show the variance of the adaptation curves across tasks and seeds (1 standard deviation). \textbf{(b)} An example adaptation behavior of Belief and PEARL agents in the sparse 2d navigation domain. The red dot represents the target the agent searches for. Different colors correspond to different episodes during adaptation.
    }
    \label{fig:thompson_vs_bayes}
\end{figure}

We further evaluate our algorithm on a suite of four meta-RL environments.
Three of our environments are modeled after common meta-RL testbeds \citep{finn2017maml, gupta2018meta}: \textbf{Cheetah velocity} in which a simulated cheetah \cite{tassa2018deepmind} has to run at an unknown speed sampled at the beginning of every episode, and the \textbf{Semicircle} environment 
with the point mass replaced with either a simulated 3-DoF rolling ball robot or a 12-DoF quadruped body (see Appendix G for details of the rolling ball body; and  Fig.~\ref{fig:pearl_gts_trajectories} for a typical trajectory of an optimal agent in this environment).
Unlike in previous papers, we train with sparse rewards only which makes these tasks considerably more difficult. The fourth environment is a novel meta-RL problem which we call \textbf{Noisy target}: the rolling ball robot is guided towards an unobserved goal location in a square with Bernoulli distributed noisy rewards which are awarded on every second time step, and whose probability of being 1 is $p = \min(0.5, (1 + d^2)^{-1}),$ where $d$ is the current distance between the agent and the target. This environment is challenging because the agent needs to filter out the uncertainty in rewards even just to estimate its distance from the target, let alone the position of the target.
Our results are summarized in Fig.~\ref{fig:further_results}.
The belief network agent supervised with the task description outperformed all agents which did not have any supervision in all environments.

\subsection{Scaling up to a complex continuous control environment}
\label{sec:numpad}
As a final testbed, we show that we can scale our approach to a complex continuous control environment which we call \textbf{Numpad}. This environment is much more difficult than standard meta-RL environments, including the ones in the previous sections, because the agent has to remember salient events which occurred more than 100 steps ago, and the tasks in $\mathcal{W}$ are POMDPs, not MDPs. We use this environment to investigate the implications of supervising with auxiliary targets other than the task description, and we also compare to shaping the agent's representation with an auxiliary head (Fig.~\ref{fig:architectures}C) instead of using a belief network.

A task in this environment requires the rolling ball to visit tiles arranged in a 3x3 grid on a platform in a prescribed sequence (which is the task description). The agent receives a +1 reward the first time it successfully completes a subsequence in the unknown sequence. Since it does not observe the task sequence, it has to remember both the longest successful subsequence, and failed attempts to extend this subsequence to act optimally. The agent also observes the tiles lighting up as long as it is stepping on them in the correct sequence. The lights reset when the ball lands on a tile which is out of sequence. Once it discovers the sequence, it can keep repeating it and collecting rewards until it runs out of time. We train the agent with a partial task cue by sampling a random mask over the task sequence at the beginning of every episode and allowing the agent to observe this masked task description. The optimal behavior is simpler when part of the task is observed, and so this extra signal helps the agent to learn e.g. useful movement primitives which can transfer to the optimal solution when the task is fully unobserved. During evaluation, the agent does not observe any part of the task.

\begin{figure}
    \centering
    \includegraphics[width=.9\linewidth]{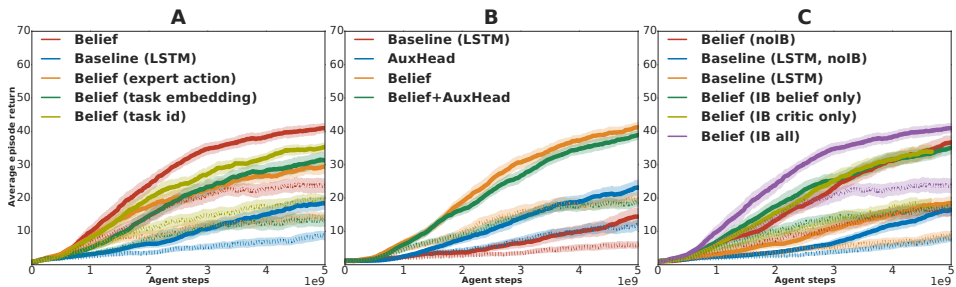}
    \caption{
    {A.} Comparison of supervision with different task-related auxiliary targets. {B.} Comparison of the belief network agent to an agent with an auxiliary head. {C.} Ablation analysis of the role of IB regularization. Solid curves correspond to performance on training tasks, while dashed curves to performance on validation tasks.}
    \label{fig:numpad_results}
\end{figure}

\begin{figure}
    \centering
    \hfill\hfill
    \begin{subfigure}[b]{0.45\linewidth}
        \centering
        \includegraphics[width=\linewidth]{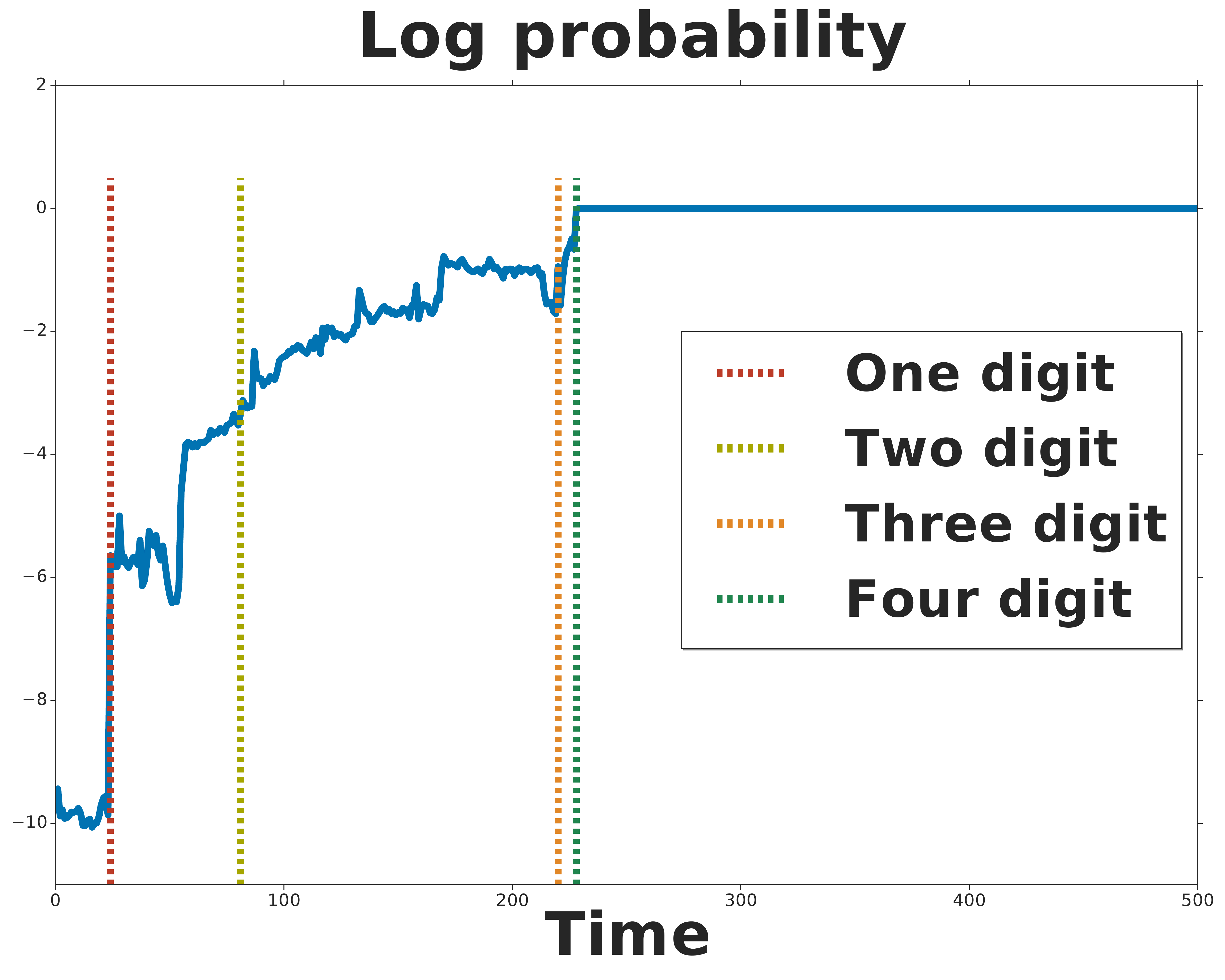}
        \caption{\label{fig:log_prob}}
    \end{subfigure}
    \hfill
    \begin{subfigure}[b]{0.45\linewidth}
        \centering
        \includegraphics[width=\linewidth]{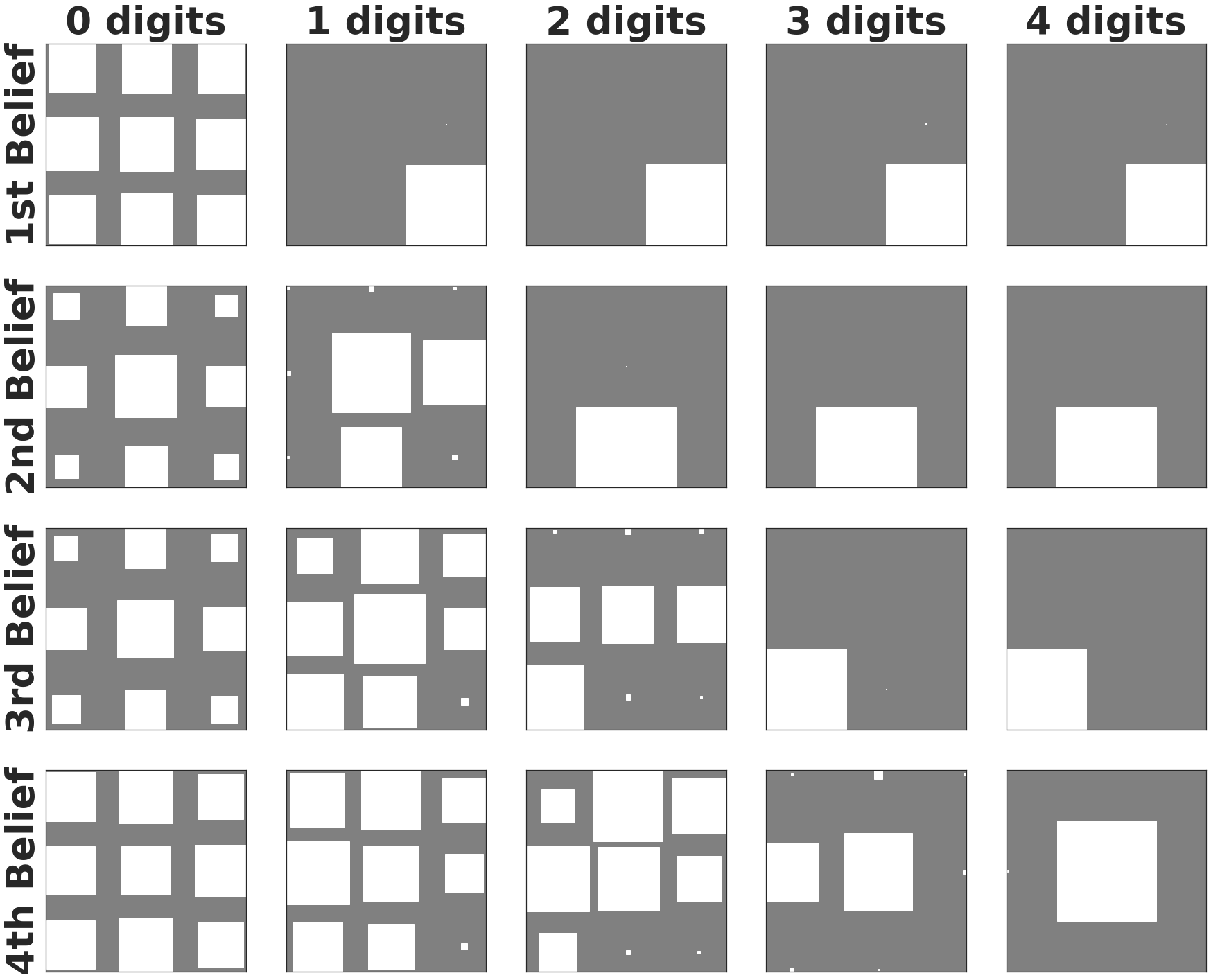}
        \caption{\label{fig:hinton_diagram}}
    \end{subfigure}
    \hfill\hfill
    \caption{\textbf{(a)} The likelihood that the agent assigns to the true task sequence during an episode rapidly increases once it discovers a new tile in the sequence. \textbf{(b)}
    Hinton diagrams visualizing beliefs about a 4 digit sequence. Each row shows the marginal probabilities for each digit. We visualize these marginals at times (columns) in an episode just before the agent discovers a new digit in the unknown task sequence (the last one is after discovering all digits). The belief of this agent reflects the contiguous structure of the allowed sequences: for example, in 3rd column, knowing that the first tile is in the lower left corner (1st row) and the second is at the center on the bottom (2nd row) makes the agent infer that the third tile (3rd row) is one of the tiles which neighbor these two.
    }
    \label{fig:visualization}
\end{figure}

There are 704 possible sequences, and we use 90\% of those as training tasks and the rest for validation. Fig. \ref{fig:numpad_results}A shows that while supervising with the task description works the best, expert actions, training task ID, and a pretrained task embedding also provide useful supervisory signal. However, using expert actions or a pretrained task embedding as a prediction target is more prone to overfitting to training tasks than supervising with the task description. In general, the observation that even with $\sim 600$ tasks we are unable to fully generalize to validation tasks is an indication that this environment is much more difficult than the previous ones in which it was possible to generalize from less than $100$ tasks. Fig.~\ref{fig:numpad_results}B shows the superiority of the belief network agent over the auxiliary head architecture. All of our agents use IB regularization, and in Fig.~\ref{fig:numpad_results}C we present an ablation analysis of the role of IB which proved essential for efficiently solving this environment. 

\subsection{Behavioral analysis}
\label{sec:behavioral_analysis}
An advantage of having access to the belief state is that one can use it to introspect and visualize what the agent is ``thinking". We show one such visualization of the evolution of an optimal agent's belief about the task during an episode in the Numpad environment in Fig.~\ref{fig:log_prob}~and~\ref{fig:hinton_diagram}.

In Fig.~\ref{fig:thompson_vs_bayes} we also compare the behavior of our agent (which is optimal) to that of an agent which tries to search using the supboptimal ``Thompson sampling" strategy, i.e. by sampling a possible task (or latent variable that represents a task) and committing to the optimal policy for this task for several steps \citep{rakelly2019efficient, gupta2018meta}. Specifically, we compare the adaptation behavior of our Belief agent to that of a PEARL agent trained using the code released as part of \cite{rakelly2019efficient} on the same sparse 2d navigation task which was used in the paper. Fig.~\ref{fig:pearl_vs_belief_adaptation} shows that our agent adapts much faster. The reason is depicted in Fig.~\ref{fig:pearl_gts_trajectories}: the Belief agent explores as much of the semicircle as possible in every episode until it finds the rewarding target without exploring parts of the semicircle which were already visited in previous episodes. In contrast, the PEARL agent simply tries a single target on the semicircle in every episode, and repeats this sampling until it hits the actual rewarding target.

\section{Discussion}

We have shown that privileged information about the unobserved task which is naturally available in meta-RL settings can be leveraged to improve the performance of both on-policy and off-policy meta-RL algorithms. Combining task-related privileged information with off-policy algorithms such as SVG(0) or SAC \citep{haarnoja2018sac} is particularly appealing because the belief state can be easily estimated from off-policy trajectories in an unbiased fashion paving a path towards highly efficient meta-RL algorithms. 
In most of our experiments, regularizing recurrent architectures with information bottleneck led to more efficient off-policy learning. We have argued that in the case of the belief network trained via supervised learning, IB helps to mitigate overfitting due to a mismatch between the empirical distribution of the replay buffer, and the trajectory distribution induced by the current policy. We leave more thorough investigation into the relationship between IB and generalization in off-policy learning to future work.

Our experiments suggest that estimating the belief state by learning the posterior of a structured task description such as an instruction or a goal location works better than learning the posterior over training task ID as well as other auxiliary targets. In the case of other auxiliary targets, this could be e.g. because our parameterization of the posterior distribution is not expressive enough, but the superiority over predicting unstructured task ID suggests that the natural structure present in the task description facilitates learning.

At last, we have also shown how the Bayes optimal adaptation strategy our agents are trained for differs from suboptimal exploration strategies based on Thompson sampling which are often studied in recent meta-RL literature.

\bibliographystyle{unsrt} %icml2019_style/icml2019}
\bibliography{literature}

\begin{appendices}
\section{Task posterior and belief-MDP derivation}
\label{sec:belief_mdp}

In this appendix we will derive the fact that the meta-RL POMDP belief state is independent of the policy given the trajectory, and that, together with the current state, it forms a sufficient statistics for computing the optimal next action.

Recall the objective being optimised by our meta-RL POMDP agent:
\begin{equation} \label{eq:main2}
    \max_{\pi} \sum_{w} p(w) \sum_{\tau_{0:\infty}} p_{\pi}(\tau_{0:\infty}|w)\left[\sum_{t=0}^{\infty} \gamma^t r_t\right].
\end{equation}
This can be rewritten as:
\begin{equation} \label{eq:main_belief}
	 \max_{\pi} \sum_{\tau_{0:\infty}} p_{\pi}(\tau_{0:\infty}) \left[\sum_{t=0}^{\infty} \gamma^t \bar{r}(x_t, a_t, x_{t+1}, b_t)\right],
\end{equation}
where 
\begin{align}
p_{\pi}(\tau_{0:\infty}) &= \sum_w p(w) p_{\pi}(\tau_{0:\infty}|w),
\\
\bar{r}(x_t, a_t, x_{t+1}, b_t) &\equiv 
	\sum_w b_t(w) \sum_r R(r|x_t, a_t, x_{t+1}, w)r, \\
b_t(w) &\equiv p(w|\tau_{0:t}),
\end{align}
are the marginal distribution of the trajectory, the posterior expected reward, and the belief state/posterior distribution over tasks, respectively.

Next, we claim that 
the posterior belief $b_t$ is independent of the policy $\pi$ given trajectory $\tau_{0:t}$, and the pair $(x_t, b_t)$ is Markovian with transition law:
\begin{align}     p(x_{t+1}|x_{t}, a_{t}, b_{t}) &= \sum_w b_{t}(w) P^w(x_{t+1}|x_{t}, a_{t}), \\
b_{t+1}(w) &= \frac{R^w(r_{t}|x_{t}, a_{t})P^w(x_{t+1}|x_{t}, a_{t}) b_t(w)}{\sum_w R^w(r_{t}|x_{t}, a_{t})P^w(x_{t+1}|x_{t}, a_{t}) b_t(w)}\quad \forall w.
\end{align}
The first statement is easy to see.  For the second statement, first we show that the belief $b_{t+1}(w) \equiv p(w|\tau_{0:t+1})$ only depends on the current and previous states, previous action, previous reward and previous belief:

\begin{align*}
    b_{t+1}(w) &\equiv p(w|\tau_{0:t+1}) = p(w|\tau_{0:t}, a_t, r_t, x_{t+1}) \\
    &=\frac{p(w, a_t, r_t, x_{t+1}|\tau_{0:t})}{\sum_w p(w, a_t, r_t, x_{t+1}|\tau_{0:t})} \\
    &= \frac{p(a_t, r_t, x_{t+1}|w, \tau_{0:t})p(w|\tau_{0:t})}{\sum_w p(a_t, r_t, x_{t+1}|w, \tau_{0:t})p(w|\tau_{0:t})} \\
    &= \frac{p(a_t, r_t, x_{t+1}|w, \tau_{0:t})b_t(w)}{\sum_w p(a_t, r_t, x_{t+1}|w, \tau_{0:t})b_t(w)} \\
\intertext{Next, since the policy $\pi(a_t|\tau_{0:t})$ is independent of the task $w$, we have that the belief is independent of the policy as well given the trajectory:}
    &= \frac{R^w(r_{t}|x_{t}, a_{t})P^w(x_{t+1}|x_{t}, a_{t})\pi(a_t|\tau_{0:t})b_t(w)}{\sum_w R^w(r_{t}|x_{t}, a_{t})P^w(x_{t+1}|x_{t}, a_{t})\pi(a_t|\tau_{0:t})b_t(w)} \\
    &= \frac{R^w(r_{t}|x_{t}, a_{t})P^w(x_{t+1}|x_{t}, a_{t})b_t(w)}{\sum_w R^w(r_{t}|x_{t}, a_{t})P^w(x_{t+1}|x_{t}, a_{t})b_t(w)},
\end{align*}
Expanding the beliefs recursively, we obtain:
\begin{align*}
    b_t(w) = p(w|\tau_{0:t}) \propto
    p(w)p_0(x_0|w)\prod_{t'=0}^{t-1} P^w(x_{t'+1}|x_{t'},a_{t'})R^w(r_{t'}|x_{t'},a_{t'},x_{t'+1}).
\end{align*}

Note that the belief state is a deterministic function of the past.
The Markovian nature of $(x_t, b_t)$ implies that in order to solve Eq.~\ref{eq:main_belief} we can restrict ourselves to policies which only depend on these variables, i.e. $\pi(a_t|\tau_{0;t}) = \pi(a_t|x_t, b_t)$.

\section{Learning task embedding}
\label{sec:task_embeddings}
In the \emph{Numpad} environment, we use a pretrained task embedding as the auxiliary target for the belief network. We learned this task embedding by jointly training a multitask policy on all \emph{training} tasks while providing a one-hot encoding of the task ID as an input. Crucially, the one-hot encoding is separately embedded via a 2-layer MLP followed by a stochastic IB layer. The columns of the output stochastic layer corresponding to a particular task ID then form a task embedding.
%that may be more structured than the task ID itself.

\section{Regularization via information bottleneck}
\label{sec:appendix:IB}
The deep variational information bottleneck \cite{alemi2017deep} regularizes neural networks by introducing a stochastic encoding $Z$ of the input $X$ as well as an additional regularization term in the objective the goal of which is to minimize the mutual information $I(Z,X)$.

In the supervised setting where we want to learn a mapping $g_{\theta}: X \rightarrow Y$ to minimize
\begin{equation*}
L(\theta, \mathcal{D}) = \EE_{(x,y) \sim \mathcal{D}} [ l(y, f_\theta(x))],
\end{equation*}
IB regularization works by introducing a latent embedding $Z$, and parameterizing $g_{\theta}$ as 
\begin{equation*}
    g_{\theta}(x) = \EE_{z \sim q_{\theta}(\cdot|x)} [f_{\theta}(z)],
\end{equation*}
where $q_\theta(z|x)$ is a stochastic encoder. The regularized objective is then to minimize the loss 
\begin{equation*}
    L_{\text{IB}}(\theta, \mathcal{D}) = \EE_{(x,y)\sim \mathcal{D}} [ \EE_{z \sim q_\theta(\cdot|x)} [ l(y, f_\theta(z))] + \lambda I[Z,X].
\end{equation*}
Although $I[Z,X]$ is intractable in general it can be upper bounded and estimated from data $\mathcal{D}$ effectively:
\begin{equation}
    \label{eq:ib_bound}
    I[Z,X] < \EE_{x \sim \mathcal{D}}[\KL[ q_{\theta}(\cdot|x) || r ]].
\end{equation}
Here, $r(z)$ is an arbitrary distribution which, in practice, is either set fixed or optimized to minimize the upper bound. Below we set $r = N(0,1)$.

While the information bottleneck regularization in \cite{alemi2017deep} was derived for supervised learning, we also regularize the policy and critic networks using a stochastic encoder and the above KL regularization even though they are trained to optimize reinforcement learning losses (see Algorithm~\ref{alg:svg0}).

\section{Results preprocessing}
\subsection{Single-threaded experiments (Section 3.1)}
Reported learning curves are the mean episodic return across 100 episodes in bandit experiments, and 20 episodes in semicircle experiments, which are smoothed with a sliding window of size 5. Each experiment is repeated 15 times in bandit experiments, and 8 times in semicircle experiments. Error bars are standard deviations of the above smoothed curves.

\subsection{Distributed setting experiments (Section 3.1 and 3.3)}
Reported learning curves represent mean episodic return averaged over distributed agents and reported on every learner iteration. The x-axis corresponds to the number of transitions processed by the learner. Curves are smoothed with a sliding window spanning 50 iterations. Each experiment is repeated 3 times, and error bars are standard deviations of the above smoothed curves.

\section{Algorithmic details for SVG(0)}
\label{sec:svg}

\subsection*{Distributed SVG(0)}

\begin{algorithm}[tb]
   \caption{Belief net SVG(0) with IB (learner)}
   \label{alg:svg0}
\begin{algorithmic}
   \STATE initial states: $h_0^{\pi}$, $h_0^{Q}$, $h_0^{b}$
   \STATE policy: $q_{t, \theta}^{\pi} \equiv q_{\theta}^{\pi}(z_t|\tau_{0:t}, h_0^{\pi})$, $\pi_{t, \theta} \equiv \pi_{\theta}(a_t|z_t)$
   \STATE Q: $q_{t, \psi}^{Q}(a_t) \equiv q_{\psi}^{Q}(z_t|\tau_{0:t}, a_t, h_0^{Q})$, $Q_{t, \psi} \equiv Q_{\psi}(z_t)$
   \STATE belief: $q_{t, \phi}^{b} \equiv q_{\phi}^{b}(z_t|\tau_{0:t}, h_0^{b})$, $b_{t, \phi} \equiv b_{\phi}(w|z_t)$
   \STATE online parameters: $\theta_O$, $\psi_O$
   \STATE target parameters: $\theta_T = \theta_O$, $\psi_T = \psi_O$
   \STATE replay buffer: $\mathcal{B}$
   \STATE batch size: $B$ 
   \STATE unroll length: $U$
   \STATE target update period: $M$
   \FOR{$step=1$ {\bfseries to} $\infty$}
   \STATE $\{h_0^{i, \pi}, h_0^{i, Q}, h_0^{i, b}, \tau_{0:U}^i, w^i \}_{i=1}^B\leftarrow$ $B$ samples from $\mathcal{R}$
   \STATE $(x_0^i, a_0^i, r_0^i, \ldots, r_{U-1}^i, x_U^i) \leftarrow \tau_{0:U}^i$
   \STATE $L_{\pi}=0, L_{Q}=0, L_{b}=0$
   \FOR{$i=1$ {\bfseries to} $B$}
   \STATE $L_{t, b}^i = \EE_{z \sim q_{t, \phi}^b} [\log b(w^i|z)] + \lambda \KL [q_{t, \phi}^{b}||N(0, 1)]$
   \STATE $(f_0^i, \ldots, f_{U-1}^i) \leftarrow q_{t, \phi}^b $ features
   \STATE $\tau_{0:U}^i \leftarrow \tau_{0:U}^i \cup (f_0^i, \ldots, f_{U-1}^i)$
   \STATE $V_t^i = \EE_{z^{\pi}\sim q_{t, \theta_T}^{\pi}} \EE_{a\sim \pi_{\theta_T}(\cdot|z^{\pi})} \EE_{z^Q\sim q_{t, \psi_T}^{Q}(a)} \left[ Q_{\psi_T}(z^Q) \right]$
   \STATE $Q_t^i = r_t^i + \gamma V_{t+1}^i$
   \STATE $Q_t^{R,i} = Retrace(Q_t^i)$
   \STATE $L_{t,Q}^i = \EE_{z\sim q_{t, \psi_O}^Q(a_t^i)}\left[(Q_{\psi_O}(z) - Q_t^{R,i})^2\right]$
   \STATE $L_{t,Q}^i \pluseq \lambda \KL [q_{t, \psi_O}^Q(a_t^i)||N(0, 1)]$
   \STATE $L_{t, \pi}^i = \EE_{z^{\pi}\sim q_{t, \theta_O}^{\pi}} \EE_{a\sim \pi_{\theta_O}(\cdot|z^{\pi})} \EE_{z^Q\sim q_{t, \psi_T}^{Q}(a)} \left[ Q_{\psi_T}(z^Q) \right]$
   \STATE $L_{t,\pi}^i \pluseq \lambda \KL [q_{t, \theta_O}^{\pi}||N(0, 1)] - \alpha \Ent [\pi_{t, \theta_O}]$
   \STATE $L_Q \pluseq \frac{1}{B} \sum_{t=0}^{U-1} L_{t, Q}^i$
   \STATE $L_{\pi} \pluseq \frac{1}{B} \sum_{t=0}^{U-1} L_{t, \pi}^i$
   \STATE $L_{b} \pluseq \frac{1}{B} \sum_{t=0}^{U-1} L_{t, b}^i$
   \ENDFOR
   \STATE $\theta_O = Adam[\nabla_{\theta_O} L_{\pi}]$
   \STATE $\psi_O = Adam[\nabla_{\psi_O} L_{Q}]$
   \STATE $\phi = Adam[\nabla_{\phi} L_{b}]$
   \IF{$ step\ \%\ M = 0$}
   \STATE $\theta_T = \theta_O$
   \STATE $\psi_T = \psi_O$
   \ENDIF
   \ENDFOR
\end{algorithmic}
\end{algorithm}
In experiments in Section 3.2 and 3.3, we use a distributed setting similar to~\citet{riedmiller2018learning}.
Several worker processes independently collect trajectories of length \textit{unroll length} of the agent's interactions with the environment, and send them to a shared replay buffer with capacity $1e6$ trajectories. A learner process (see Algorithm~\ref{alg:svg0}) then uniformly samples batches of trajectories from the replay buffer, updates the networks via a gradient descent step on the appropriate SVG(0) losses augmented with the auxiliary belief loss from Section 2.1, KL regularization terms from Eq.~\ref{eq:ib_bound}, and policy entropy regularization. The learner then shares the updated network parameters with the workers. We store internal states of all RNNs in the replay buffer together with the observations and use them for initializing these RNNs during training. Agents which utilize information bottleneck can either always sample from the stochastic encoder, or only sample during training and use the mean of the encoder's distribution during acting. We typically tried different settings and chose the one which worked the best. Instead of using KL regularization in information bottleneck, we also experimented with fixing the variance of the layer and tuning the variance instead of the KL regularization coefficient.

\subsection*{Single-threaded SVG(0)}
In Section 3.1 we use a single-threaded implementation instead of our distributed setup. A single update to the networks is the same as described in Algorithm~\ref{alg:svg0} except we use TD(0) instead of Retrace to estimate targets for critic training, and we interleave network updates and data collection as follows: We collect a trajectory of length \textit{unroll length}, and then we do \textit{n\_train\_iters} actor and critic network updates and one belief network update.

\subsection*{Network architectures}
Actor, critic, and belief networks all encode inputs with a 2 or 3-layer MLP with ELU activation functions except for the first layer where we apply layer normalization \cite{leiba2016layernorm} followed by a TANH activation (no layer norm in single-threaded experiments). In the critic network we augment the encoded inputs with the action (processed with a TANH) to be evaluated. The networks then pass the encoded inputs to a LSTM network. If we include IB regularization, then the LSTM outputs are linearly projected to the parameters of a diagonal Gaussian distribution. Actor and critic networks linearly map either samples from the IB encoder, or outputs of the LSTM to either parameters of the policy distribution or scalar value function. Belief network maps the LSTM outputs (or IB encoder samples) to parameters of a distribution over the auxiliary target using another 1 or 2-layer MLP with ELU activations. The head architecture processes the actor and critic's LSTM outputs (or IB encoder samples) with a 2-layer MLP. We parameterize the policy and IB encoder with parameters $(\mu_{a}, \log\sigma_{a})$ which are mapped to the means and standard deviations of a diagonal Gaussian distribution $\mathcal{N}(\tilde{\mu_{a}}, \tilde{\sigma_{a})}$ using the mapping: 
\begin{equation*}
    \tilde{\mu_{a}} = \tanh(\mu_{a}),
\end{equation*}
\begin{equation*}
    \tilde{\sigma_{a}} = 0.001 + (1.0 - 0.001) \sigma (\log\sigma_{a}),
\end{equation*}
where $\sigma$ is a sigmoid function.

All networks are optimized with the Adam optimizer with default TensorFlow settings except the learning rate.

Parameterizations of the belief distributions are listed in Appendix G for each environment.

\subsection*{Hyperparameters--distributed setting}

We use the following default hyperparameters in our experiments which we further tune for each environment via a combination of grid search and informal search over learning rates, entropy bonus, IB coefficients or variance in IB layers, and size of the bottleneck layer.

\textbf{Default hyperparameters}\newline
\textit{Actor learning rate}, $0.0003$ \newline
\textit{Critic learning rate}, $0.0003$ \newline
\textit{Belief network learning rate}, $0.0003$ \newline
\textit{Target update period}: $500$ \newline
\textit{Actor network encoder}: MLP with sizes $(200, 200, 200)$ \newline
\textit{Critic network encoder}: MLP with sizes $(300, 300, 300)$ \newline
\textit{Actor LSTM size}: $100$ \newline
\textit{Critic LSTM size}: $300$ \newline
\textit{Belief LSTM size}: $100$ \newline
\textit{Batch size}: $512$ \newline
\textit{Unroll length}: $50$ \newline
\textit{Entropy bonus}: $0.001$ \newline
\textit{Discount factor}: $0.99$ \newline
\textit{Number of parallel actors}: $200$ \newline
\textbf{Belief bottleneck parameters}\newline
\textit{Belief bottleneck dimension}: $20$ \newline
\textit{Belief bottleneck loss coefficient}: $0.1$ \newline

\textbf{Agent bottleneck parameters (Belief agents)}\newline
\textit{Actor bottleneck dimension}: $20$ \newline
\textit{Actor bottleneck loss coefficient}: $0.01$ \newline
\textit{Critic bottleneck dimension}: $20$ \newline
\textit{Critic bottleneck loss coefficient}: $1.0$ \newline

\textbf{Agent bottleneck parameters (LSTM agents)}\newline
\textit{Actor bottleneck dimension}: $100$ \newline
\textit{Actor bottleneck loss coefficient}: $0.01$ \newline
\textit{Critic bottleneck dimension}: $200$ \newline
\textit{Critic bottleneck loss coefficient}: $1.0$ \newline

\textbf{Task embedding parameters}\newline
\textit{Embedding size}: $100$ \newline
\textbf{Auxiliary loss parameters}\newline
\textit{Head dimensions}: MLP with sizes $(200, 200)$ \newline

\subsection*{Hyperparameters--single process}

We use the following hyperparameters which were chosen using a combination of grid search and informal search over learning rates, IB coefficients, and entropy bonus.

\textbf{Hyperparameters}\newline
\textit{Actor learning rate}, $0.0005$ (MLP), $0.00005$ (LSTM) \newline
\textit{Critic learning rate}, $0.0005$ (MLP), $0.00005$ (LSTM) \newline
\textit{Belief network learning rate}, $0.0005$ \newline
\textit{Target update period}: $500$ \newline
\textit{Actor network encoder}: MLP with sizes $(256, 256)$ \newline
\textit{Critic network encoder}: MLP with sizes $(256, 256)$ \newline
\textit{Actor LSTM size}: $128$ \newline
\textit{Critic LSTM size}: $128$ \newline
\textit{Belief LSTM size}: $128$ \newline
\textit{Batch size}: $100$ \newline
\textit{Unroll length}: $10$ \newline
\textit{Entropy bonus}: $0.01$ \newline
\textit{Discount factor}: $0.99$ \newline
\textbf{Belief bottleneck parameters}\newline
\textit{Belief bottleneck dimension}: $16$ \newline
\textit{Belief bottleneck loss coefficient}: $1.$ \newline
\textit{Actor bottleneck dimension}: no bottleneck \newline
\textit{Critic bottleneck dimension}: $128$ \newline
\textit{Critic bottleneck loss coefficient}: $0.01$ \newline

\section{Algorithmic details for PPO}
\label{sec:ppo}
All architectures are the same as in single-thread SVG(0) except in bandit experiments where encoders are $(128, 128)$ and LSTM uses layer norm. While the policy loss is clipped according to the PPO objective from \cite{schulman2017proximal}, the value function and belief network are updated using the same number of gradient descent steps as the policy without any clipping. In bandit experiments, the belief network agent excludes belief features from the value network which is always recurrent because we find them unnecessary.

Hyperparameters are listed below. In bandit experiments, we searched over belief network learning rate, discount, generalized advantage lambda, entropy coefficient, and number of gradient steps per iteration. In semicircle experiments, we searched over learning rates, and entropy coefficient. 

\textbf{Hyperparameters--Bandit}\newline
\textit{Actor learning rate}: $0.0005$ \newline
\textit{Value learning rate}: $0.001$ \newline
\textit{Belief network learning rate}: $0.003$ \newline
\textit{Discount}: $0.99$ \newline
\textit{Entropy}: $0.05$ \newline
\textit{Generalized advantage lambda}: $0.3$ \newline
\textit{Batch size}: $10000$ \newline
\textit{Gradient descent steps per update}: $10$ \newline
\textit{Epsilon in PPO clipped objective}: $0.2$ \newline
\textit{Belief loss weight in Head architecture}: $10.$ \newline

\textbf{Hyperparameters--Semicircle}\newline
\textit{Actor learning rate}: $0.0005$ \newline
\textit{Value learning rate}: $0.0005$ \newline
\textit{Belief network learning rate}: $0.0005$ \newline
\textit{Discount}: $0.99$ \newline
\textit{Entropy}: $0.01$ \newline
\textit{Generalized advantage lambda}: $0.95$ \newline
\textit{Batch size}: $2000$ \newline
\textit{Gradient descent steps per update}: $4$ \newline
\textit{Epsilon in PPO clipped objective}: $0.2$ \newline
 \newline

\section{Implementation details of our environments}
\label{sec:implementation_details}

In this section we describe our environments in more detail. Most meta-RL papers assume a K-shot formulation in which the agent is given $K$ episodes of length $T$ to explore and adapt. We do not explicitly assume such K-episode formulation. Instead, we will only have a single episode, and the $K$-shot structure will be present implicitly in the dynamics of the environment, for example by having the environment reset to the initial state after reaching a goal state.

\subsection{Point mass semicircle}
The point mass is specified by its position and orientation in a plane which also form an observation for the agent. Action space is two-dimensional corresponding to moving forward/backward with a given velocity, and rotating with a given angular velocity. Maximum speed is $1$, and maximum angular velocity is $2\pi$. One timestep is $0.01$, and we use an action repeat of $10$. The targets are distributed on a semi-circle of radius $0.2$. After reaching the target, the agent is awarded a +1 reward and the point mass is teleported to the origin. At the beginning of each episode, the point mass starts at the origin with a random orientation. One episode is $100$ action steps long. The task description is one-dimensional corresponding to an angle on the semi-circle. Belief distribution is parameterized as a piece-wise constant function with 10 constant intervals.

\subsection{Multi-armed bandit}
On every step, the agent pulls one of $K$ arms, and obtains a random reward drawn from a Bernoulli distribution with success probability $p_i$, where $i$ is the arm number. The goal of the agent is to maximize the total reward collected during $T$ pulls without knowing what the arm probabilities $p_i$ are. The task description is a vector of arm probabilities, and the action space is discrete corresponding to the $K$ arms. There are no observations but as in all other environments the agent's input contains the action taken and reward obtained on the previous step. We choose the task distribution to be the uniform distribution on the $K$-dimensional unit hypercube. Belief distribution is parameterized using a Beta distribution.

\subsection{Cheetah}
A simulated half-cheetah \citep{tassa2018deepmind} is supposed to run at an unknown speed sampled uniformly from the interval $[3.0, 10.0]$. On every step, the agent receives a reward
\begin{equation*}
    r(v) = \text{max}\left(1 - \left|\frac{v}{v_{\text{target}}} - 1\right|, 0\right),
\end{equation*}
where $v$ is the cheetah's current speed, and $v_{\text{target}}$ is the target speed which is also the task description. One episode is 10 seconds long, and consists of $1000$ steps. Belief distribution is parameterized using a Gaussian distribution.

\subsection{Rolling ball/quadruped semi-circle}
The environment is implemented using the MuJoCo simulator \citep{todorov2012mujoco} and the high-level logic is the same as in the previous semicircle environment except that the point mass is replaced with either a rolling ball robot or 12 DoF quadruped. The rolling ball possesses a 3 dimensional continuous action space and moves by applying torque in order to rotate around the z-axis, or to accelerate in the forward direction. It can also jump by actuating an invisible slide joint, although the tasks that we consider do not require jumping. The observations consist of the robot's position and orientation on the platform and several proprioceptive features such as joint positions and velocities which are necessary for movement control. Each episode consist of 1000 steps. Belief distribution is parameterized using a Gaussian distribution.

\subsection{Noisy target}
We use the same rolling ball as in the previous environment. Unlike in semicircle, the ball is not teleported to origin upon reaching the target, so the difficulty of this environment stems from having to integrate noisy rewards rather then from remembering where the target it. Belief distribution is parameterized using a Gaussian distribution.

\subsection{Numpad}

\begin{figure}[!htb]
    \centering
    \includegraphics[width=\linewidth]{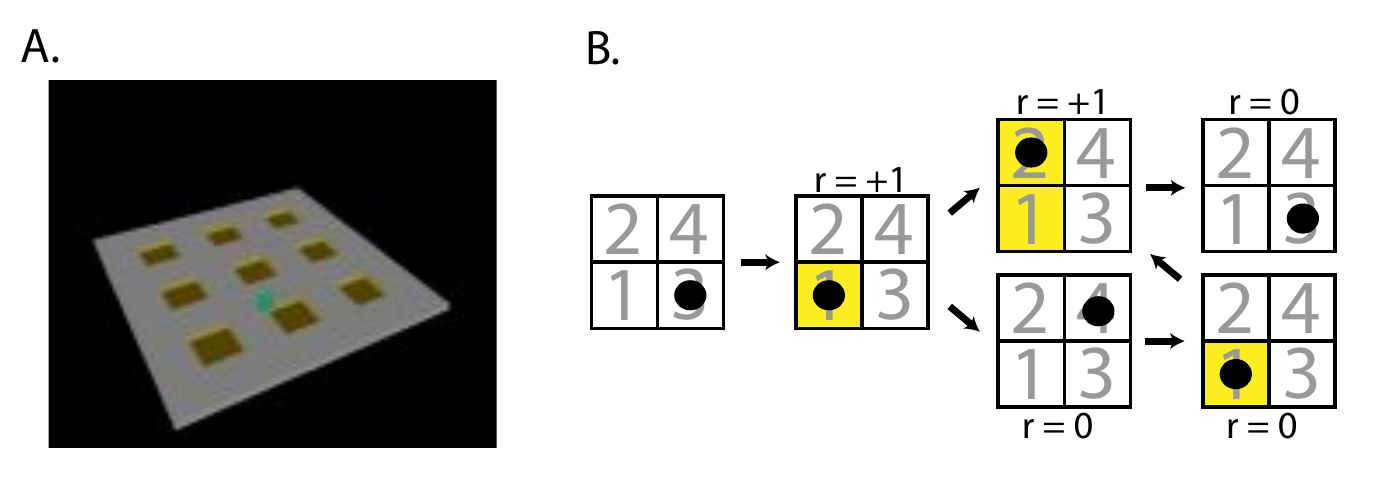}
    \caption{A. Numpad environment. B. The logic of the tasks. We show a simplified example of transitions on a 2x2 grid. In the actual environment, the grid is 3x3, and the robot must learn to move from tile to tile by actuating the robot shown in A.}
    \label{fig:numpad_env}
\end{figure}

The task is for the rolling ball robot to complete a sequence of movements between tiles arranged in a 3x3 grid on a two-dimensional platform. See Fig.~\ref{fig:numpad_env}. A task consists of visiting tiles in a prescribed sequence. As long as the agent visits tiles in the right order, the tiles light up, but if it touches a tile out of sequence, the lights get reset, and the agent needs to start the sequence from the beginning. If the sequences are of length 4, then the goal is to get to a state where 4 tiles are activated. The activation pattern of the tiles is part of the observation. To improve exploration during learning the agent is gradually guided towards the correct sequence by a +1 reward that is given the first time it completes a valid subsequence. Therefore, it has to estimate its belief about the true sequence by remembering both the longest successful subsequence, and failed attempts to extend this subsequence. When the sequence is completed for the first time, the rewards and lights are reset, and it can keep collecting rewards until it runs out of time.

We note that the reward function is history dependent, and so the environment is partially observed even when the task is known.

We restrict ourselves to contiguous sequences of neighboring tiles of length at most 4. The task description $w$ is a sequence of four numbers from the set $\{-1, 1, 2, \dots 9 \}$, where we use $-1$ as a placeholder if the sequence is shorter than $4$. In total, there are $704$ tasks. Distributions over the task description are parameterized as four independent categorical distributions assigning probabilities to each digit. The random mask which forms the task cue observed by the agent during training is sampled by first uniformly sampling from the set $\{0, 1, 2, 3, 4\}$ to obtain the number of digits which will be hidden from the agent, and then uniformly sampling from the set of all binary masks with that number of zeros.

\section{Generalization across tasks}
\label{sec:task_genearlization}
To investigate generalization across tasks, we ran some of our experiments with differently sized task training sets. Our results are summarized in this section. Our figure descriptions differ slightly from the main text: Belief is equivalent to Belief (noIB), Belief + IB is equivalent to Belief, LSTM is equivalent to Baseline (LSTM, noIB), and LSTM + IB is equivalent to Baseline (LSTM).

\subsection{Multi-armed bandit}

\begin{figure}[!htb]
    \centering
    \includegraphics[width=\linewidth]{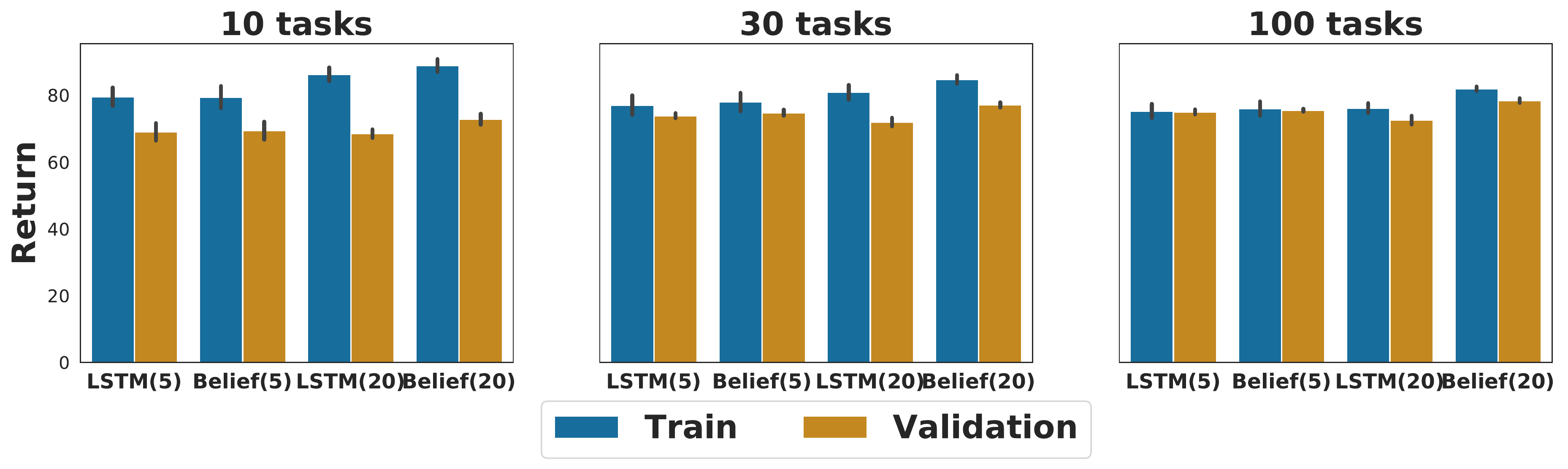}
    \caption{Training vs. validation performance of the baseline LSTM agent and the best performing agent on multi-armed bandit problems. Agents are evaluated after 500 iterations of training. The number in parentheses indicates the number of arms.}
    \label{fig:bandits_generalization}
\end{figure}

Fig.~\ref{fig:bandits_generalization} compares training and validation performance after 500 iterations of training in the bandit environment. In addition to the 20-armed environment from main text, we also show results for 5 arms.

\subsection{Quadruped semi-circle}

\begin{figure}[!htb]
    \centering
    \includegraphics[width=\linewidth]{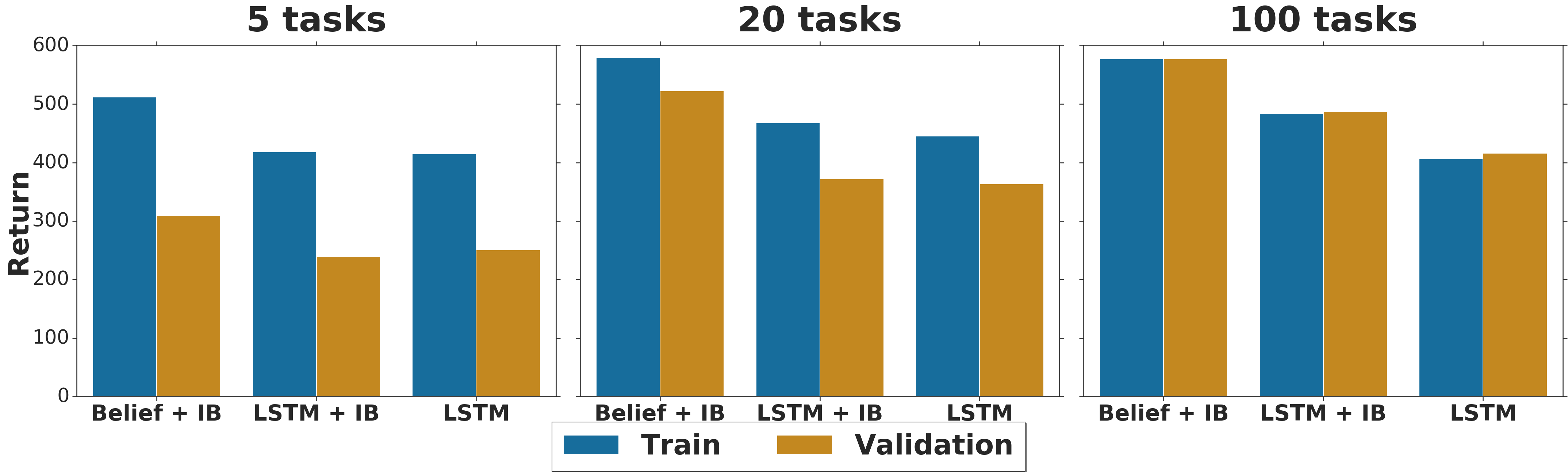}
    \caption{Dependence of the generalization gap on the number of training tasks in the \emph{Quadruped semi-circle} environment. We show the performance after $5e9$ learner updates.}
    \label{fig:final_gotosphere_train_size}
\end{figure}
\begin{figure}[!htp]
    \centering
    \includegraphics[width=0.8\linewidth]{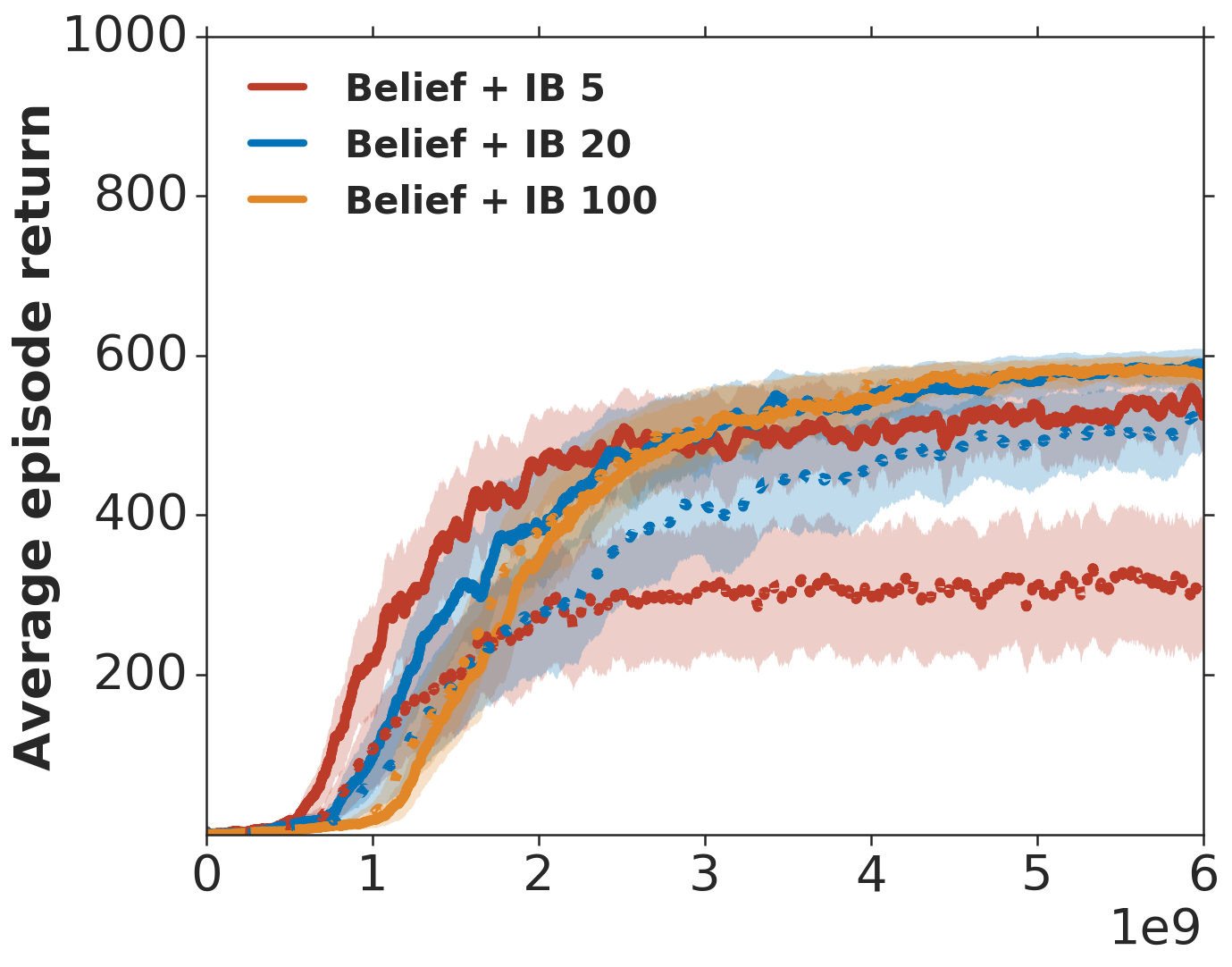}
    \caption{Training/validation learning curves of the belief architecture with IB regularization in the \emph{Ant semicircle} environment for various training set sizes.}
    \label{fig:gts_train_size_unsup}
\end{figure}

Fig.~\ref{fig:final_gotosphere_train_size} shows the training set size dependence of the generalization gap and Fig.~\ref{fig:gts_train_size_unsup} shows full training/validation learning curves for various training set sizes.

\subsection{Numpad}
\begin{figure}[!htb]
    \centering
    \includegraphics[width=\linewidth]{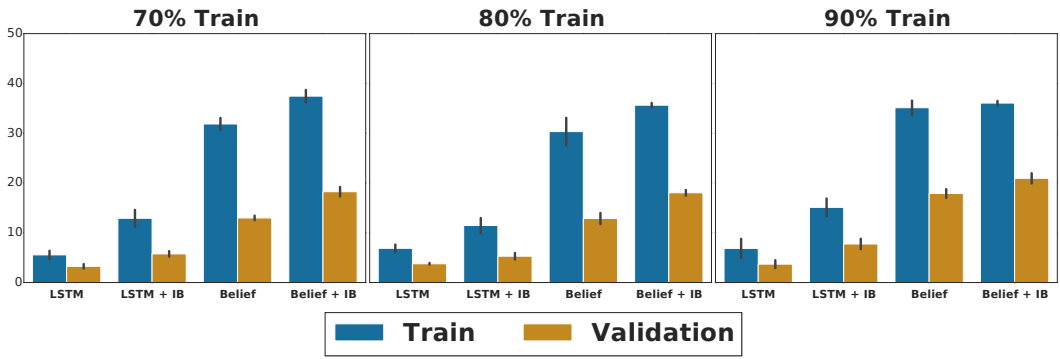}
    \caption{Dependence of the generalization gap on the number of training tasks in the \emph{Numpad} environment. Percentages correspond to fractions of all possible tasks. We show performance after $4e9$ learner updates.}
    \label{fig:final_numpad_train_size_bar}
\end{figure}
\begin{figure}[!htb]
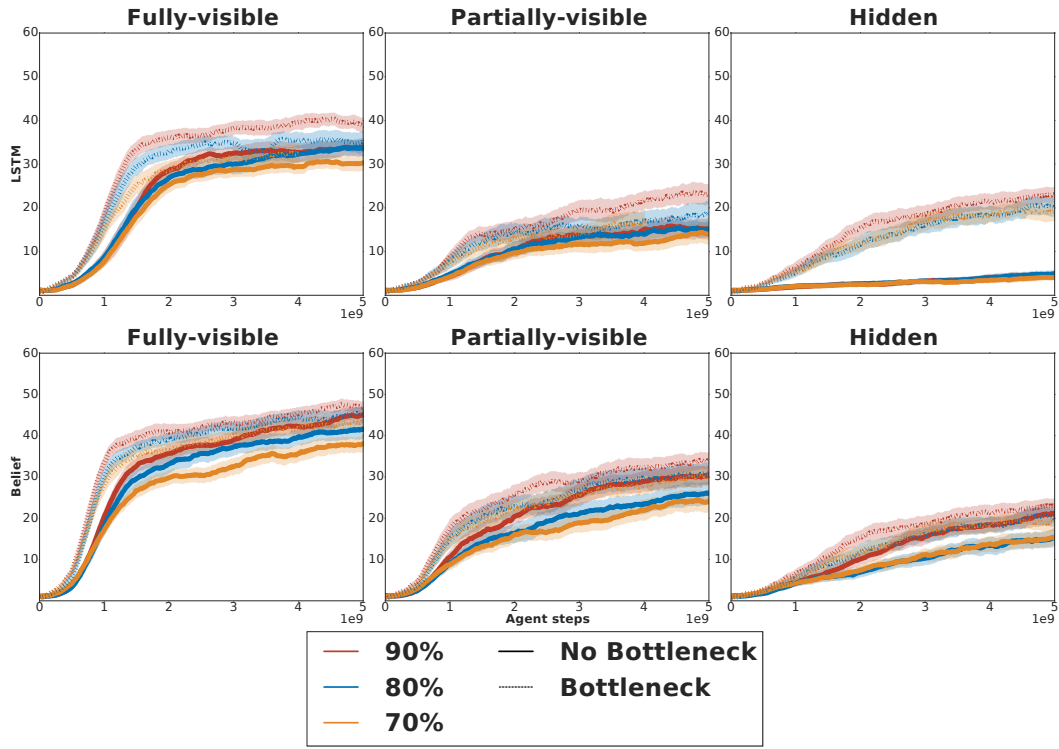

    \centering
    \includegraphics[width=\linewidth]{./figures_arxiv/final_numpad_generalisation_gap_final_unsup.pdf}
    \includegraphics[width=\linewidth]{./figures_arxiv/final_numpad_generalisation_gap_final_sup.pdf}
    \caption{Training/validation learning curves of the baseline LSTM architecture (\textbf{upper curve}), and the belief agent (\textbf{bottom curve}) with and without information bottleneck regularization in the \emph{Numpad} environment. Percentages correspond to fractions of all possible tasks.}
    \label{fig:numpad_validation_size}
\end{figure}

In the Numpad environment the agent observes randomly masked task description during training. We are mainly interested in the agent's performance when it does not observe any part of the task, which is also what we report in the main text and by default also here. However, it is also insightful to look at the performance of the agent on episodes in which it fully observes the task description, or in which it partially observes the task description (the training regime). We will refer to these different evaluation regimes as \emph{fully visible} when the agent sees the task, \emph{partially visible} when the task is partially masked out, and \emph{fully hidden} when the agent does not observe any part of the task (default evaluation regime).

Fig.~\ref{fig:final_numpad_train_size_bar} shows the training set size dependence of the generalization gap. Fig.~\ref{fig:numpad_validation_size} shows full training/validation learning curves for various training set sizes.

% \bibliography{literature}
% \bibliographystyle{icml2019_style/icml2019}
\end{appendices}

\end{document}